\documentclass[journal]{IEEEtranTIE}
\usepackage{graphicx}
\usepackage{cite}
\usepackage{picinpar}
\usepackage{amsmath}
\usepackage{amsthm}
\usepackage{url}
\usepackage{flushend}
\usepackage[latin1]{inputenc}
\usepackage{colortbl}
\usepackage{soul}
\usepackage{multirow}
\usepackage{pifont}
\usepackage{color}
\usepackage{alltt}
\usepackage[hidelinks]{hyperref}
\usepackage{threeparttable}
\usepackage{enumerate}
\usepackage{siunitx}
\usepackage{breakurl}
\usepackage{epstopdf}
\usepackage{pbox}
\usepackage{pifont}

\usepackage[comma,numbers,square,sort&compress]{natbib}
\usepackage{subfigure}
\usepackage{tikz}
\usetikzlibrary{positioning}

\theoremstyle{plain}

\newtheorem{assumption1}{Assumption}

\newtheorem{theorem1}{Theorem}

\newtheorem{remark1}{Remark}

\theoremstyle{definition}

\newtheorem{proof1}{Proof}

\begin{document}
\title{Active Admittance Control with Iterative Learning for General-Purpose Contact-Rich Manipulation}

\author{
	\vskip 1em
	
	Bo Zhou,
	Yuyao Sun,
	Wenbo Liu,
	Ruixuan Jiao,
	\\ Fang Fang,
	and Shihua Li

	\thanks{
	
		This work was supported in part by the National Natural Science Foundation (NNSF) of China under Grant 62073075.
		
		B. Zhou, Y. Sun, W. Liu, R. Jiao, F. Fang, and S. Li are with the Key Laboratory of Measurement and Control of CSE(School of Automation, Southeast University), Ministry of Education, Nanjing 210096, China (e-mail: {zhoubo@seu.edu.cn}, {220211791@seu.edu.cn}, {lwb@seu.edu.cn}, {220232086@seu.edu.cn}, {ffang@seu.edu.cn}, {lsh@seu.edu.cn}).
	}
}

\maketitle

\begin{abstract}
Force interaction is inevitable when robots face multiple operation scenarios. How to make the robot competent in force control for generalized operations such as multi-tasks still remains a challenging problem. Aiming at the reproducibility of interaction tasks and the lack of a generalized force control framework for multi-task scenarios, this paper proposes a novel hybrid control framework based on active admittance control with iterative learning parameters-tunning mechanism. The method adopts admittance control as the underlying algorithm to ensure flexibility, and iterative learning as the high-level algorithm to regulate the parameters of the admittance model. The whole algorithm has flexibility and learning ability, which is capable of achieving the goal of excellent versatility. Four representative interactive robot manipulation tasks are chosen to investigate the consistency and generalisability of the proposed method. Experiments are designed to verify the effectiveness of the whole framework, and an average of 98.21\% and 91.52\% improvement of RMSE is obtained relative to the traditional admittance control as well as the model-free adaptive control, respectively.
\end{abstract}


{}

\definecolor{limegreen}{rgb}{0.2, 0.8, 0.2}
\definecolor{forestgreen}{rgb}{0.13, 0.55, 0.13}
\definecolor{greenhtml}{rgb}{0.0, 0.5, 0.0}

\section{Introduction}

\begin{figure}[!htb]
\centering
\includegraphics[width=8.8cm]{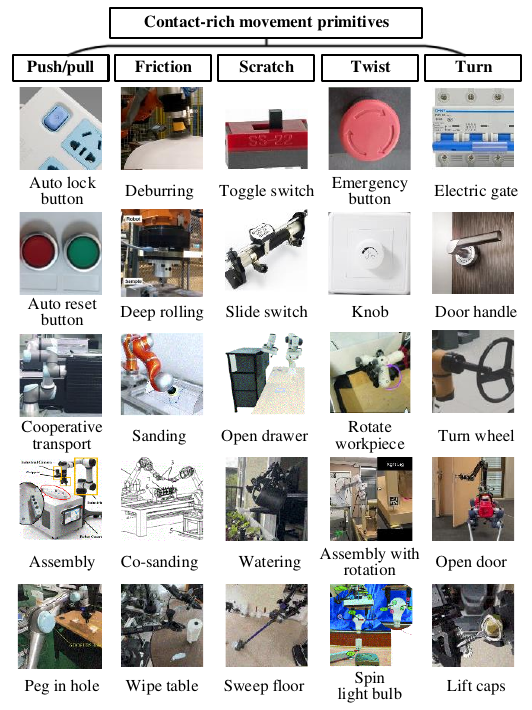}
\caption{Contact-rich movement primitives categorized into five classes, each of these contains multiple operations or multiple operation objects of the same class}
\label{all_tasks}
\end{figure}

\IEEEPARstart {W}{ith} the development of robotics technology, the application scenarios of various types of robots are gradually increasing. In addition to the industrial scenario applications mainly based on constant force control, there are many daily life scenario applications mainly based on variable force control. In this paper, the applications in each scenario are categorized into five major categories of operation primitives with force, as shown in the Fig \ref{all_tasks}. For the industrial production field, here are sanding, polishing, deep rolling, spraying, large object handling, assembly, etc. In the field of daily life, robots need to realize more interaction tasks than in the industrial field in order to assist or even integrate into human life. For robots, it is still a challenging problem to realize these force interaction applications using a unified framework.

For areas such as unified frameworks and multi-tasks, the methods used by researchers are generally dominated at the perception and planning level, while force control is rarely addressed in the known scope. Some researchers have used methods based on reinforcement learning and transfer learning to realize a single task by dividing the primitives, trajectory learning, and so on. Whereas, if migrating to multi-task, the means of migration learning and imitation learning are frequently used for expansion. Although the means of perception and position control can solve many interaction problems without force control, there are still a large number of interaction problems containing force control that cannot be solved.

Impedance control proposed by Hogan, as the most widely used force control method, can obtain good results when applied to robots. However, impedance control has three parameters and requires a reference position and force to be adjusted. Therefore, for multi-tasks, not only the reference need to be modified, but also the parameters need to be adjusted. Even for the tracking of the same curve, the parameters of the initial and subsequent parts need to be changed for better tracking and robustness.

Most of the existing methods for force interaction of robotic have been developed within the framework of impedance and admittance. The development paths are mainly divided into two types, the improvement of the parameters, and the improvement of the reference position and force. For the first type, some of these parameters are usually adjusted by adaptive and fuzzy methods to make them more capable of force control tasks. However, this type of improvement is unable to learn and is mostly aimed at constant force tasks. Therefore, it usually has better results after discarding the initial oscillations. For the second type, it is usually performed in an iterative learning manner. The inputs are adjusted from generation to generation so that it performs well from start to finish in a single iteration. This type of methods will show increasingly accurate control in a single task, but once transferred to other tasks, the force control parameters have to be readjusted. Most of the above force control methods are multi-task unfriendly, so a highly generalizable force control framework is urgently needed to be proposed.

The main contributions of this paper include:

(1) A highly generalizable and learnable interaction force control framework is proposed for multi-task scenarios with repeatability. Observing common multi-task operations, they tend to have repeatability. And the need for manual parameter tuning limits the application of the basic force control in multi-tasking. In contrast, the iterative learning strategy adopted in this paper ensures the generalization of the control due to its use of a model-free approach.

(2) The ILC-MBK method is proposed for parameter tuning of the admittance model. First, this paper modifies the original admittance model by transforming the parameters m, b, and k and using them as system inputs. The modified model is more convenient for parameter tuning and application in iterative learning. Then, this paper establishes the ILC-MBK method by inverse modeling to explore the relationship between the parameters and the tracking performance, and then the update law of parameters. Meanwhile, this paper proves the convergence on the infinite paradigm through the error transfer matrix.

(3) Interactive force control experiments for multi-tasks are applied and validated. To the best of the authors' knowledge, this paper is the first multi-task oriented research on interaction force control. The ILC-MBK method in this paper obtains good results in the operation tasks of three selected typical switches. It has better results, as well as sustained convergence, compared to the CFDL-MFAC method. Meanwhile, experiments also verify the high generalizability and success rate of the method.

\section{Related Work}
\subsection{Multi-task Manipulation}
In the field of multi-task operation, researchers usually use reinforcement learning, imitation learning, and behavioral cloning to learn the trajectories of multiple tasks\cite{Park2023Research}. Liu \textit{et al.}\cite{Liu2020Reinforcement} used reinforcement learning for the skill learning problem so that the robotic arm learns the fetch and place positions, and then realizes the fetching and placing of objects. Wang \textit{et al.}\cite{Wang2020Meta} used meta-learning to realize the ant robot walking in multiple specified scenarios in a simulation environment. Cheng \textit{et al.}\cite{Cheng2023Multi} used multi-task reinforcement learning to achieve better operation results for multiple tasks in a meta-world simulation environment. Arcari \textit{et al.}\cite{Arcari2023Bayesian} proposed a Bayesian multi-task learning based robot operation MPC scheme, which solves the control problem under the dynamic uncertainty of the system. Wu \textit{et al.}\cite{Wu2022Framework} and Bai \textit{et al.}\cite{Bai2022Robot} used imitation learning for the scenarios of three-button operation and wheel-type valve operation, respectively, to complete the operation while relying on the position judgment only. Yang \textit{et al.}\cite{Yang2018Interface} used a human-robot interaction system to enable a human to operate the robot to accomplish a task, which was transformed into a robot's skill. Andrew \textit{et al.}\cite{Silva2022LanCon} proposed LanCon-Learn, which combines the attention mechanism and reinforcement Learning, which combines attention mechanisms and reinforcement learning to enable it to accomplish multiple tasks in both simulation and real species, but it requires linguistic means to guide the robot to switch to a new task. From the above work, it can be seen that the current field of multitasking is dominated by learning tools, and most of them control position or trajectory with little or no consideration of force interaction.

\subsection{Interactive Force Control Methods}
Hogan\cite{Hogan1984Impedance} proposed impedance control to mimic a mass-spring damping system between the robot's end and the environment, which in turn gave the robot a means of force control that would be widely used in future research. Raibert \textit{et al.}\cite{Raibert1981Hybrid} were the first to propose a feedback-based linearized hybrid position/force control algorithm and developed two complementary orthogonal subspaces to decouple position and force control. Anderson \textit{et al.}\cite{Anderson1988Hybrid} proposed hybrid impedance control, which combines impedance control with hybrid position/force control. Ikeura \textit{et al.}\cite{Ikeura1995Variable} proposed variable impedance control in which the impedance depends on the robot speed. The above traditional force control methods, with excellent force control performance, can only be used as the underlying framework, but cannot be applied individually to high-demand applications, such as constant force control and variable force control, due to the fact that they require input from the reference force position and the existence of certain controller parameters that need to be adjusted. Wahrburg \textit{et al.}\cite{Wahrburg2016MPC} combined MPC with conductive control to address the instability of conductive control when in contact with the outside world. Fukui \textit{et al.}\cite{Fukui2018Design} introduced the estimation of the stiffness or impedance of the interaction environment into iterative learning of force control. Norouzi \textit{et al.}\cite{Norouzi2019Robotic} introduced PD control and fuzzy gain into ILC learning to help tracking tasks. Huynh \textit{et al.}\cite{Huynh2019Force/Position} and Chen \textit{et al.}\cite{Chen2020Robotic} developed a gradient-based ILC control algorithm, which improves the force control accuracy of the robotic arm, retains the advantages of suppleness and low cost, and is applied to applications such as parallel robots and robotic roll pressing, respectively. Ren \textit{et al.}\cite{Ren2021Learning}, which combines the RBFNN with the sliding-mode control, enables the robot to have the ability of force control for soft environments such as surgery. Iskandar \textit{et al.}\cite{Iskandar2023Hybrid} developed an extended Cartesian impedance control algorithm, which realizes explicit force tracking through geometric constraints with a hybrid approach to give it a high force control performance in sanding. The above method combines the impedance-conductance control and force-position hybrid control at the bottom layer with various kinds of regulation methods at the top layer, so that the overall better control effect is achieved. Among the above methods, the use of ILC's can make the system achieve good results at the beginning rather than after the force control is stabilized, so ILC will be the main focus in the subsequent research.
\subsection{Iterative Learning Methods}
Iterative learning has undergone considerable development since Arimoto\cite{Arimoto1984Bettering} proposed iterative learning control. Adaptive ILC was proposed by French \textit{et al.}\cite{French2000Nonlinear} to solve the ILC problem with a standard Lyapunov design, using a standard adaptive controller and starting parameter estimation with the final values obtained in the previous iteration. Later on, model-free adaptive control was proposed by Hou\cite{Hou2011Data}, which enables the overall control framework to accomplish proper control by linearization even in the absence of a model. Hui \textit{et al.}\cite{Hui2021Extended} combined the model-free adaptive ILC with an expanding state observer, which gives it a perturbation resistance capability. Lin \textit{et al.}\cite{Lin2021Event} combined model-free adaptive ILC with an event-triggered nonlinear system and extended it accordingly. In the subsequent development, Bu \textit{et al.}\cite{Bu2022Event} and Yu \textit{et al.}\cite{Yu2021Data} integrated the model-free adaptive ILC into a systematic approach, proved its convergence by constructing a Lyapunov function, and applied it to event-triggered systems and nonlinear MIMO systems, respectively, so that the application area of the application of iterative learning to be greatly improved. Although the model-free method can have better results for traditional nonlinearities, it is more limited for systems with input and state coupling such as parameter tuning, and it is more difficult to achieve good control results. And the model-free adaptive ILC in the proof, usually with the help of the concept of spectral radius, as the lower limit of all the paradigms, although the limitation of the spectral radius can make the system error in the final convergence to zero, but sometimes cannot guarantee that its iterative learning process has been to maintain a good performance, so iterative convergence algorithms based on other paradigms urgently need to be proposed.

\section{Problem Formulation}

\subsection{Platform and Operations}
\begin{figure}[!htb]
\centering
\includegraphics[width=8.8cm]{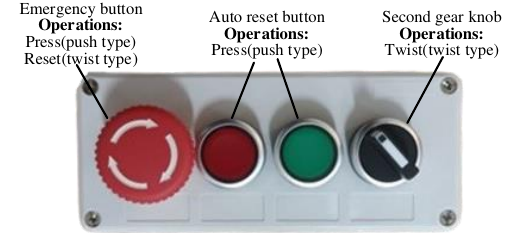}
\caption{Target switch box with three typical switches, auto reset button, second gear knob and emergency button}
\label{fig3}
\end{figure}

For multi-task interactive force control scenarios, a wide variety of operations are included. In order not to lose the generality, three typical switches are selected as research object, as shown in Fig \ref{fig3}. The switch box includes two auto reset buttons, a second gear knob, and an emergency button. It contains two of the five major types of operations, namely push-pull operation and twist operation.

\begin{figure}[!htb]
\centering
\subfigure[View 1 of the multi-functional end-effector]{
\label{fig41}
\includegraphics[width=4.2cm]{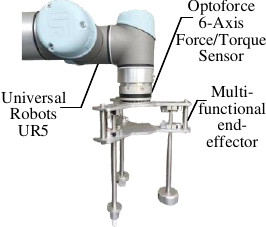}}
\subfigure[View 2 of the multi-functional end-effector]{
\label{fig42}
\includegraphics[width=4.2cm]{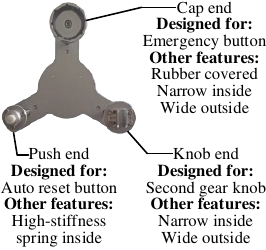}}
\caption{Hardware platform, contains Universal Robots UR5 robotic arm, OptoForce 6-Axis Force/Torque Sensor and a multi-functional end-effector}
\label{fig4}
\end{figure}

In order to interact with the switches, this paper has designed a multi-functional end-effector for the Universal Robots UR5 robotic arm and OptoForce 6-Axis Force/Torque Sensor, as shown in Fig \ref{fig4}. In view 2, at its lower left is the push end for pressing the auto reset button. On the lower right is the knob end for twisting the second gear knob. Above is a cap end for pressing and resetting the emergency button.

With the assistance of the end-effector, the UR5 arm can perform four operations: press auto reset button, twist second gear knob, press emergency button, and reset emergency button. Since the reaction force varies with position, they are regarded as rigid-based manipulation objects in this paper.

\begin{figure}[!htb]
\centering
\subfigure[Auto reset button]{
\label{figd1}
\includegraphics[width=2.0cm]{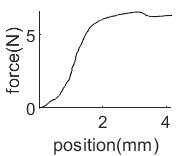}}
\subfigure[Second gear knob]{
\label{figd2}
\includegraphics[width=2.0cm]{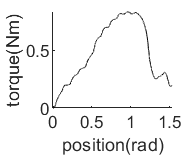}}
\subfigure[Emergency button-press]{
\label{figd3}
\includegraphics[width=2.0cm]{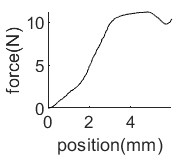}}
\subfigure[Emergency button-reset]{
\label{figd4}
\includegraphics[width=2.0cm]{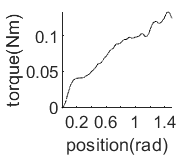}}
\caption{Filtered force-position relation, respectively pressing auto reset button, twisting second gear knob, pressing emergency button and resetting emergency button}
\label{datas}
\end{figure}

In order to obtain the position force characteristics of the switches, this paper accomplishes the operations by manually controlling the UR5 arm. With the help of fine-tuned filters, flatter position force curves are obtained. Since the operation of the robotic arm is somewhat different from that of a human, some fluctuation data are retained in the data acquisition to demonstrate the generalization of tracking. The sampled position force curves are shown in Fig \ref{datas}. This confirms the rigidity-dominated characteristics of each object. It also shows that the rigidity exhibited at different operation stages is different from the initial one. This suggests the need for more sophisticated modeling in system construction, or the estimation of stiffness drived by feedback data. Subsequent chapters will keep approaching this position force curve as a learning objective.

\section{Manipulation System Design}
In the field of multi-task operation targeted in this paper, most of the operations are repeatable. And since the parameters of force control need to be different among multi-tasks, iterative learning will be extremely suitable. Therefore, the traditional admittance control model is no longer used. In this part, a force control model with mbk parameters and their deformations as inputs is first constructed based on the admittance control model, and then the ILC-MBK algorithm is constructed to accomplish the specified multi-tasks and a convergence proof is given.

\subsection{Modeling of admittance towards parameter tuning}

Considering the position control widely used in real robots, this paper adopts the admittance control with position as output. Its basic equations are as follows:
\begin{equation}
m\Delta \ddot x+b\Delta \dot x+k\Delta x=\Delta f\label{eq0}
\end{equation}

Where $m$, $b$, and $k$ are the inertia, damping and stiffness of the system respectively, $\Delta \ddot x$, $\Delta \dot x$, $\Delta x$ are the errors of acceleration, velocity and position respectively, $\Delta f$ is the error of the interaction force between the robot and the environment.

When applied to the actual multi-task interaction operation, the admittance control has certain application difficulties, i.e., it is not generalized to different task parameters and requires a front-loaded parameter adjustment link. In order to adapt to the multi-task interactive operation with variable parameters, in this paper, the $m, b, k$ parameters of admittance control are deformed and regarded as system inputs. The equation \ref{eq0} is directly transformed to:
\begin{align}
\nonumber \Delta \ddot x(t)=-\frac{b(t)}{m(t)}(\dot x_r(t)-\dot x(t))-\frac{k(t)}{m(t)}(x_r(t)-x(t)) \\
+\frac{1}{m(t)}(f_r(t)-f(t))\label{eq1}
\end{align}

The discrete equations can be written as:
\begin{align}
\Delta \dot x(t+\Delta t)&=(\dot x_r(t)-\dot x(t))+\Delta \ddot x(t)\Delta t \\
\Delta x(t+\Delta t)&=(x_r(t)-x(t))+\Delta \dot x(t+\Delta t)\Delta t \\
\dot x(t)&=x_r(t)-\Delta \dot x(t) \\
x(t)&=x_r(t)-\Delta x(t) \\
f(t)&=k_{env}(t)x(t)\label{eq2}
\end{align}

The $k_{env}(t)$ means the rigidity of environment or the operation object. Replace the $\dot x(t)$ with $v(t)$, and reform equation (\ref{eq1})-(\ref{eq2}) into discrete system model:
\begin{align}
\nonumber \begin{bmatrix}\Delta v(t+\Delta t) \\ \Delta x(t+\Delta t) \end{bmatrix}=\begin{bmatrix}1-\frac{b(t)}{m(t)}\Delta t & -\frac{k(t)}{m(t)}\Delta t & \frac{1}{m(t)}\Delta t \\ \Delta t-\frac{b(t)}{m(t)}\Delta t^2 & 1-\frac{k(t)}{m(t)}\Delta t^2 & \frac{1}{m(t)}\Delta t^2 \end{bmatrix} \\
\nonumber \left( \begin{bmatrix}v_r(t) \\ x_r(t) \\ f_r(t) \end{bmatrix}-\begin{bmatrix}v(t) \\ x(t) \\ f(t) \end{bmatrix}\right)\\
\begin{bmatrix}v(t) \\ x(t) \\ f(t) \end{bmatrix}=\begin{bmatrix}1 & 0 & 0 \\ 0 & 1 & 0 \\ 0 & k_{env}(t) & 0 \end{bmatrix}\begin{bmatrix}v_r(t) \\ x_r(t) \\ f_r(t) \end{bmatrix}-\begin{bmatrix}1 & 0 \\ 0 & 1 \\ 0 & k_{env}(t) \end{bmatrix}\begin{bmatrix}\Delta v(t) \\ \Delta x(t) \end{bmatrix}\label{eq3}
\end{align}

For further consideration, the system is rewritten as a discrete system with some constant matrices and vectors:
\begin{align}
\nonumber x(t+\Delta t)=&(\beta u^T (t)A+E)(r(t)-y(t))\\
y(t)=&C(t)x(t)+D(t)r(t)\label{eq4}
\end{align}

In which, $x(t)=\begin{bmatrix}\Delta x(t) & \Delta v(t) \end{bmatrix}^T$, $u(t)=\begin{bmatrix}\frac{b(t)}{m(t)} & \frac{k(t)}{m(t)} & \frac{1}{m(t)}\end{bmatrix}^T$, $y(t)=\begin{bmatrix}v(t) & x(t) & f(t)\end{bmatrix}^T$, $r(t)=\begin{bmatrix}v_r(t) & x_r(t) & f_r(t)\end{bmatrix}^T$. The constant matrices and vectors $\beta, A, E$ assist $u(t)$ to achieve the correct matrix form. $C(t), D(t)$ are related to the rigidity of the interaction object. And $D(t)$ is a linear transform of $C(t)$, $D(t)=C(t)\cdot T^D_C$. The corresponding parameter estimation for $C(t)$ is needed to achieve better control effect.

\subsection{ILC-MBK Algorithm}

\begin{figure}[!htb]
\centering
\includegraphics[width=8.8cm]{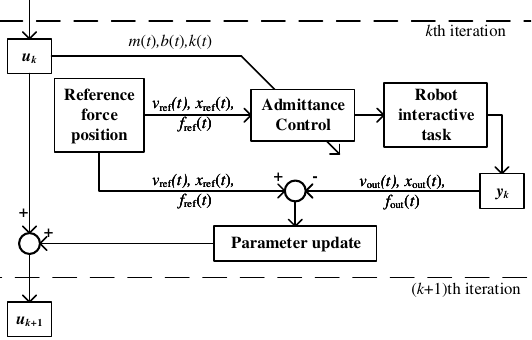}
\caption{ILC-MBK procedure during iterations, $u_k$ means the parameter input of admittance control, reference force position gives the reference to admittance control and the update law. The system output $y_k$ can be gotten after the interactive tasks, then the update law makes the $u_k$ go to the new generation $u_{k+1}$}
\label{ilc_1}
\end{figure}

Considering the learnable characteristics of ILC, this paper propose ILC-MBK algorithm, which combines ILC and admittance control to achieve the optimization of parameter. The procedure among iterations is shown as Fig.\ref{ilc_1}.

Let $U(t)=\beta u^T (t)A+E$, considering the system model without any correction:
\begin{align}
\nonumber y(t+\Delta t)=C(t+\Delta t)&U(t)(r(t)-y(t)) \\
+&D(t+\Delta t)r(t+\Delta t)\label{eq5}
\end{align}

Construct a correction term $\Delta U(t)$, assume the addition of $\Delta U(t)$ makes the output approach the reference values. Therefore, construct the error of output $e(t+\Delta t)=r(t+\Delta t)-y(t+\Delta t)$, and the system model with correction is:
\begin{align}
\nonumber y(t+\Delta t)+e(t+\Delta t)&=C(t+\Delta t)(U(t)+\Delta U(t)) \\
&e(t)+D(t+\Delta t)r(t+\Delta t)\label{eq6}
\end{align}

The differences of equation (\ref{eq5}) and (\ref{eq6}) can be gotten:
\begin{align}
e(t+\Delta t)=C(t+\Delta t)\Delta U(t)e(t)\label{eq7}
\end{align}

With the help of matrix generalized inverse, the following equation is obtained:
\begin{align}
\Delta U(t)=C^+(t+\Delta t)e(t+\Delta t)e^+(t)\label{eq8}
\end{align}

Due to the progressive approach of ILC, equation (\ref{eq8}) should be multiplied by the learning rate $\alpha$:
\begin{align}
\alpha \Delta U(t)=\alpha C^+(t+\Delta t)e(t+\Delta t)e^+(t)\label{eq9}
\end{align}

Thus, a part of the ILC-MBK is proposed as following:
\begin{align}
\nonumber U_{k+1}(t)=&U_k (t)+\alpha \Delta U_k(t)\\
=&U_k (t)+\alpha C^+_k(t+\Delta t)e_k(t+\Delta t)e^+_k(t)\label{eq10}
\end{align}

Considering the uncertain value $C(t)$, the estimation matrix $\hat{C}_k (t)$ and the cost function of estimation $J(C(t))$ is constructed:
\begin{align}
\nonumber J(C(t))=(y_{k-1}(t)-&C(t)(x_{k-1}(t)+T^D_Cr(t)))^2\\
+&\mu(C(t)-\hat{C}_{k-1}(t))^2\label{eq11}
\end{align}

Let $\frac{\partial J(C(t))}{\partial C(t)}=0$, it can be gotten:
\begin{align}
\nonumber \hat{C}_k&(t)=\hat{C}_{k-1}(t)\\
\nonumber +&\frac{\eta(y_{k-1}(t)-\hat{C}_{k-1}(t)(x_{k-1}(t)+T^D_Cr(t)))}{\mu+\Vert x_{k-1}(t)+T^D_Cr(t)\Vert^2} \\
*&(x_{k-1}(t)+T^D_Cr(t))^T\label{eq12}
\end{align}

In which $\mu >0$, $0<\eta \le 2$. As a result, a whole system (\ref{eq13}) and ILC-MBK algorithm (\ref{eq14}) (\ref{eq15}) through iterations can be gotten:
\begin{align}
\nonumber x_k&(t+\Delta t)=U_k(t)(r(t)-y_k(t))\\
y_k&(t)=C(t)x_k(t)+D(t)r(t)\label{eq13}\\
U_{k+1}&(t)=U_k (t)+\alpha\hat{C}^+_k (t+\Delta t)e_k(t+\Delta t)e^+_k(t)\label{eq14}\\
\nonumber \hat{C}_k&(t)=\hat{C}_{k-1}(t)\\
\nonumber +&\frac{\eta(y_{k-1}(t)-\hat{C}_{k-1}(t)(x_{k-1}(t)+T^D_Cr(t)))}{\mu+\Vert x_{k-1}(t)+T^D_Cr(t)\Vert^2} \\
*&(x_{k-1}(t)+T^D_Cr(t))^T\label{eq15}
\end{align}
\subsection{Convergence Analysis}

For a smooth convergence proof, the following assumptions are made.

\begin{assumption1}
For system (\ref{eq13}), the initial value of system is identical over all iterations, which means $y_k (0)=y_0$ with $y_0$ being a constant.
\end{assumption1}
\begin{assumption1}
Let $t_1\ne t_2$, $\alpha_0=\alpha C(t_1)\hat{C}^+_k (t_2)$, the ILC learning rate and the infinity norm satisfy $\Vert\alpha_0\Vert_\infty<\Vert C(t+\Delta t)U_k(t)\Vert _\infty <\overline{b}$.
\end{assumption1}

\begin{remark1}
In practical systems, the above assumptions can be easily satisfied. For Assumption 1, this can be done by controlling the initial state at each iteration, which in turn makes the initial error consistent. For Assumption 2, it can be done by numerically limiting the updates at each iteration, which in turn allows the corresponding part to be controlled.
\end{remark1}

For the considered system (\ref{eq13}), assuming that the termination time of an iteration is $t_{end}$, and let $n=t_{end}/\Delta t$, the relationship of $e_{k+1}(t)$ and $e_k(t)$ among different times can be easily gotten. Discard the initial error and convert to the matrix form:
\begin{align}
e_{k+1}=Ge_k\label{eq25}
\end{align}
\begin{align}
G=\begin{bmatrix}a_n&b_{n-1}c_{n-1}&...&b_2 c_2 \displaystyle\prod_{l=3}^{n-1} d_l&b_1 c_1 \displaystyle\prod_{l=2}^{n-1} d_l\\
0&a_{n-1}&...&b_2 c_2 \displaystyle\prod_{l=3}^{n-2} d_l&b_1 c_1\displaystyle\prod_{l=2}^{n-2} d_l\\
...&...&...&...&...\\
0&0&...&a_2&b_1 c_1\\
0&0&...&0&a_1\end{bmatrix}\label{eq26}
\end{align}

Among them, $e_{k}=\begin{bmatrix}e_{k}(n\Delta t)&...&e_{k}(2\Delta t)&e_{k}(\Delta t)\end{bmatrix}^T$, $a_i=\sum_{l=0}^{i}(-\alpha_0)^l$, $b_i=\sum_{l=1}^{i}-(-\alpha_0)^l$, $c_i=C((i+1)\Delta t)U_k (i\Delta t)$, $d_i=-C((i+1)\Delta t)U_{k+1} (i\Delta t)$.
\begin{theorem1}
Considering the system (\ref{eq13}) which satisfy the Assumption 1 and 2, similarly to the above matrix $G$, the matrix can be obtained:
\begin{align}
G_m=\begin{bmatrix}\Vert a_n\Vert_\infty &...&\Vert b_2 c_2\displaystyle\prod_{l=3}^{n-1} d_l\Vert_\infty &\Vert b_1 c_1\displaystyle\prod_{l=2}^{n-1} d_l\Vert_\infty \\
0&...&\Vert b_2 c_2\displaystyle\prod_{l=3}^{n-2} d_l\Vert_\infty &\Vert b_1 c_1\displaystyle\prod_{l=2}^{n-2} d_l\Vert_\infty \\
...&...&...&...\\
0&...&\Vert a_2\Vert_\infty &\Vert b_1 c_1\Vert_\infty \\
0&...&0&\Vert a_1\Vert_\infty\end{bmatrix}\label{eq31}
\end{align}

The infinity norm of $G_m$ matrix satisfy:
\begin{align}
\Vert G\Vert_\infty\le \Vert G_m \Vert_\infty<1\label{eq32}
\end{align}
\end{theorem1}

\begin{proof1}
Let $n_\alpha=\Vert \alpha_0\Vert_\infty, n_{cu}=\Vert C(t+\Delta t)U(t)\Vert_\infty$, the infinity norm of $G_m$ satisfy:
\begin{align}
\nonumber \Vert G_m\Vert_\infty&=max^n_{i=1}(\frac{1-(-n_\alpha)^{i+1}}{1+n_\alpha}\\
\nonumber &+\frac{n_\alpha-(-n_\alpha)^i}{1+n_\alpha}n_{cu}+\cdots+\frac{n_\alpha-(-n_\alpha)^2}{1+n_\alpha}n_{cu}^{i-1}) \\
\nonumber &=\frac{1+n_\alpha \frac{n_{cu}-n_{cu}^i}{1-n_{cu}}}{1+n_\alpha}-\frac{(-n_\alpha)^{i+1}\frac{1-(-n_{cu}/n_\alpha)^i}{1+n_{cu}/n_\alpha}}{1+n_\alpha} \\
&\le \frac{1+n_\alpha\frac{n_{cu}}{1-n_{cu}}}{1+n_\alpha}<1\label{eq33}
\end{align}

Due to the properties of infinity norm,
\begin{align}
\Vert G\Vert_\infty\le \Vert G_m \Vert_\infty<1\label{eq34}
\end{align}
\qed
\end{proof1}

Construct the Lyapunov function for the $k_{th}$ iteration:
\begin{align}
V_k=\Vert e_k\Vert ^2_\infty\ge 0\label{eq35}
\end{align}

As a result,
\begin{align}
\nonumber \Delta V_{k+1}=&V_{k+1}-V_k\\
\nonumber =&\Vert Ge_k\Vert _\infty^2-\Vert e_k\Vert _\infty^2\\
\nonumber \le& (\Vert G_m\Vert _\infty^2-1)\Vert e_k\Vert _\infty^2\\
\le& 0\label{eq36}
\end{align}

Therefore the error of the system will converge to zero in the iterative domain. At the same time, the numerical limitations in Assumption 2 will have an impact on the properties of sustained convergence.

\section{Results of Experiments}

\subsection{Experiments Preparation}
Since the robot platform provides 125Hz data exchange frequency, $t=0.05s$ is chosen as the control period in this paper.

Prior to the main experiment, four positional force profiles were collected and filtered. A set of initial admittance control parameters were determined in table \ref{initp}.
\begin{table}[htbp]
\renewcommand{\arraystretch}{1.3}
\caption{Initial parameters of admittance control}
\centering
\label{initp}
\begin{tabular}{c c c c}
\hline\hline \\[-3mm]
Operation              & $m$    & $b$     & $k$     \\[1.6ex] \hline
pressing auto reset button      & $12$   & $19$    & $2784$  \\
twisting second gear knob       & $1.0$  & $0.945$ & $0.945$ \\
pressing emergency button       & $20$   & $7$     & $2840$  \\
resetting emergency button      & $0.06$ & $0.084$ & $0.084$ \\
\hline\hline
\end{tabular}
\end{table}

Meanwhile, three sets of experiments are set up for pure admittance control, CFDL-MFAC algorithm with optimizing parameters, and the ILC-MBK method in this paper. In order to show the difference in learning performance, the number of iterations is set to 750. Among them, the pure admittance control only uses the initial parameters for force control. The CFDL-MFAC algorithm\cite{Yu2021Data} was originally designed for nonlinear system to track a trajectory using a model-free adaptive approach, while the parameter tuning involved in this paper can be considered as a complex nonlinear system, and thus it is taken into account in this paper.

\subsection{Experiment Results}

\begin{figure}[!htb]
\centering
\includegraphics[width=8.8cm]{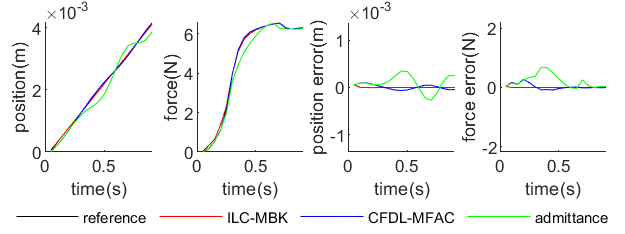}
\caption{Learning results of pressing auto reset button, shows the differences among admittance control, CFDL-MFAC, and our method ILC-MBK}
\label{sr1}
\end{figure}

\begin{figure}[!htb]
\centering
\includegraphics[width=8.8cm]{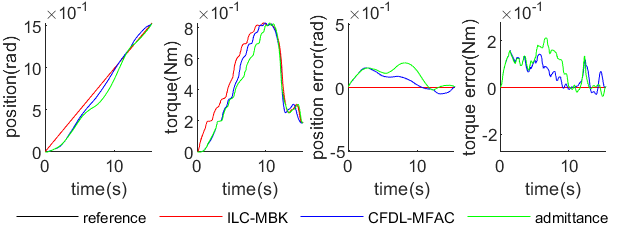}
\caption{Learning results of twisting second gear knob, shows the differences among admittance control, CFDL-MFAC, and our method ILC-MBK}
\label{sr2}
\end{figure}

\begin{figure}[!htb]
\centering
\includegraphics[width=8.8cm]{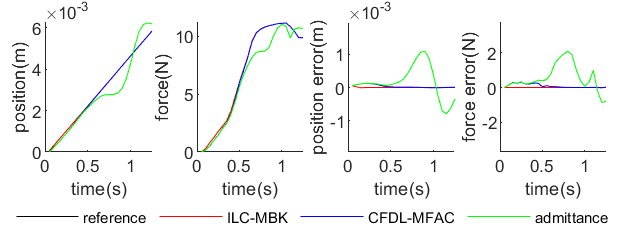}
\caption{Learning results of pressing emergency button, shows the differences among admittance control, CFDL-MFAC, and our method ILC-MBK}
\label{sr3}
\end{figure}

\begin{figure}[!htb]
\centering
\includegraphics[width=8.8cm]{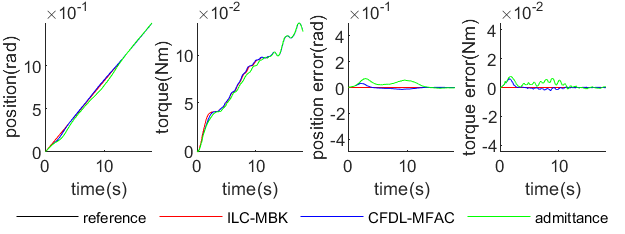}
\caption{Learning results of resetting emergency button, shows the differences among admittance control, CFDL-MFAC, and our method ILC-MBK}
\label{sr4}
\end{figure}

The results of the experiments for the four operations are shown in Fig.\ref{sr1}-\ref{sr4}. The macroscopic learning data are shown in Table \ref{table}. Where RMSE denotes the root mean square error and MDR denotes the maximum decrease rate.

\begin{table*}[!htb]
\renewcommand{\arraystretch}{1.3}
\caption{Macro data among different methods in experiments}
\centering
\label{table}
\begin{threeparttable}
\begin{tabular}{c c c c c c c}
\hline\hline \\[-3mm]
\multicolumn{2}{c}{Case}                          & Admittance RMSE          & CFDL-MFAC RMSE           & CFDL-MFAC MDR            & ILC-MBK RMSE                      & ILC-MBK MDR                       \\[1.6ex] \hline
\multirow{2}*{pressing auto reset button}&position& $1.927901\times 10^{-4}$ & $4.247734\times 10^{-5}$ & $1.489895\times 10^{-5}$ & $\mathbf{1.428727\times 10^{-5}}$ & $\mathbf{8.250754\times 10^{-5}}$ \\
~                                        &force   & $3.056389\times 10^{-1}$ & $9.556722\times 10^{-2}$ & $2.589517\times 10^{-2}$ & $\mathbf{9.009233\times 10^{-3}}$ & $\mathbf{7.929894\times 10^{-2}}$ \\ \hline
\multirow{2}*{twisting second gear knob} &position& $1.177933\times 10^{-1}$ & $8.133682\times 10^{-2}$ & $8.809201\times 10^{-4}$ & $\mathbf{3.359886\times 10^{-4}}$ & $\mathbf{1.172015\times 10^{-1}}$ \\
~                                        &torque  & $1.074198\times 10^{-1}$ & $8.047331\times 10^{-2}$ & $7.423573\times 10^{-4}$ & $\mathbf{4.433461\times 10^{-5}}$ & $\mathbf{1.067252\times 10^{-1}}$ \\ \hline
\multirow{2}*{pressing emergency button} &position& $5.107179\times 10^{-4}$ & $5.515682\times 10^{-5}$ & $1.287761\times 10^{-4}$ & $\mathbf{1.161062\times 10^{-5}}$ & $\mathbf{4.612571\times 10^{-4}}$ \\
~                                        &force   & $9.348815\times 10^{-1}$ & $1.349617\times 10^{-1}$ & $2.180634\times 10^{-1}$ & $\mathbf{6.485387\times 10^{-4}}$ & $\mathbf{7.932928\times 10^{-1}}$ \\ \hline
\multirow{2}*{resetting emergency button}&position& $3.315011\times 10^{-2}$ & $1.066907\times 10^{-2}$ & $6.054934\times 10^{-4}$ & $\mathbf{3.613088\times 10^{-5}}$ & $\mathbf{3.309176\times 10^{-2}}$ \\
~                                        &torque  & $2.970556\times 10^{-3}$ & $1.480440\times 10^{-3}$ & $4.204934\times 10^{-5}$ & $\mathbf{3.609412\times 10^{-5}}$ & $\mathbf{2.934093\times 10^{-3}}$ \\
\hline\hline
\end{tabular}
\begin{tablenotes}
\footnotesize
\item[1] Three main methods are experimented, pure admittance control, CFDL-MFAC, and our method ILC-MBK.
\item[2] Two technical indicators of learning method are considered, RMSE means Root Mean Square Error, MDR means Maximum Descent Rate.
\end{tablenotes}
\end{threeparttable}
\end{table*}

As can be seen from the graphs, the ILC-MBK method proposed in this paper has the advantage of continuous convergence and better tracking results. The pure admittance control has no learning ability, so its error is relatively large. The CFDL-MFAC algorithm is based on dynamic linearization, so according to the inverse model analysis, the input is in a complex relationship with the output, which may make the method appear a certain degree of difficulty. The CFDL-MFAC method is based on the spectral radius, which can guarantee only the convergence of transfer matrix of error, but not the continuous convergence of error.

In terms of the initial parameters of the different characteristics, the more fixed the stiffness of the characteristic, the wider the range of parameters it can accommodate. When resetting emergency button, it has many sets of parameters that can achieve better results, while when twisting second gear knob, its optional parameter range is narrower. For a target with a fixed stiffness, only a reasonable parameter value is needed to make the initial force control behave reasonably, and the subsequent force control will also behave reasonably. For targets with varying stiffness, different parameters are required for different stiffnesses, and thus different tolerance ranges, resulting in a narrower range of parameters to choose from initially.

\section{Conclusion}
Aiming at the lack of a generalized force control framework for interaction tasks, this paper proposes ILC-MBK algorithm. The admittance control is used as bottom layer to ensure flexibility, and iterative learning control is used as top layer to learn the parameters, thus realizing the purpose of generalization. In this paper, three rigid-based objects are used, and experiments with robotic arm and multiple end-effectors are designed to verify the generalization ability and success rate of the force control framework.

In this paper, when facing multiple interaction objects, admittance model parameters are chosen to be learned instead of inputs, which improves its generalization to a certain extent. When considering the force control characteristics of each object, which is mainly rigid, a local linearization with the help of a model-free method is chosen to make it closer to the real model.

The direction of future work will focus on modeling and elimination of system disturbances, sensor errors, and so on. Meanwhile, since the parameter update law in this paper is relative to errors, it also needs to be investigated in order to enhance robustness.

\bibliographystyle{IEEEtranTIE}
\bibliography{mine}

\begin{IEEEbiography}[{\includegraphics[width=1in,height=1.25in,clip,keepaspectratio]{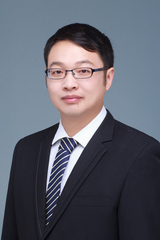}}]
{Bo Zhou} (Member, IEEE) received the B.S. degree in automation from the University of Science and Technology of China, Hefei, China, in 2003, and the Ph.D. degree in automation from Shenyang Institute of Automation, Chinese Academy of Sciences, Shenyang, China, in 2009.

He is currently an Associate Professor with the School of Automation from Southeast University, Nanjing, China. His research interests include industrial/mobile robot control technology.
\end{IEEEbiography}
\begin{IEEEbiography}[{\includegraphics[width=1in,height=1.25in,clip,keepaspectratio]{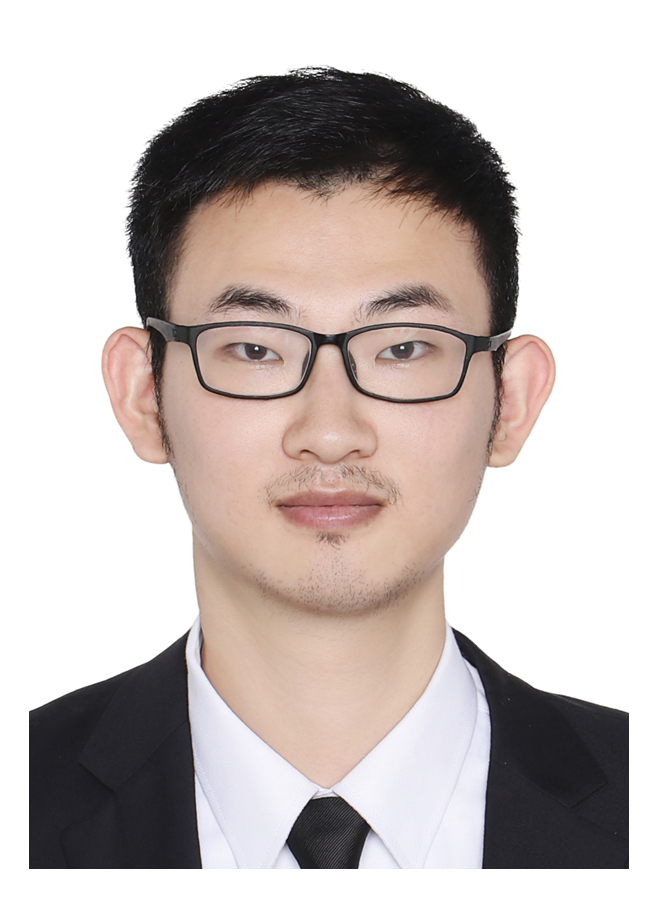}}]
{Yuyao Sun} received the B.S. degree in automation from Southeast University, Nanjing, China, in 2021. He is currently pursuing the M.S. degree with the School of Automation, Southeast University, Nanjing, China.

His main research interests include robot force control, iterative learning control, behavior tree.
\end{IEEEbiography}
\begin{IEEEbiography}[{\includegraphics[width=1in,height=1.25in,clip,keepaspectratio]{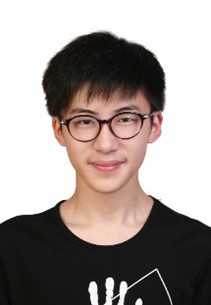}}]
{Wenbo Liu} received the B.S. degree in process equipment and control engineering from Fuzhou University, Fuzhou, China, in 2020. Currently, he is working toward the Ph.D. degree with School of Automation, Southeast University, Nanjing, China.

His main research interests include task planning, compliant control and industrial robots.
\end{IEEEbiography}
\begin{IEEEbiography}[{\includegraphics[width=1in,height=1.25in,clip,keepaspectratio]{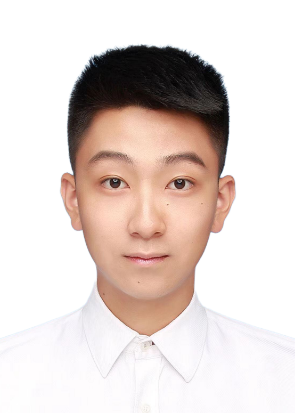}}]
{Ruixuan Jiao} received the B.S. degree in Automation
from China University of Mining and Technology, Xuzhou, China, in 2023. He is currently working toward the M.S. degree in robotic engineering with Southeast University, Nanjing, China.

His research interests include learning based planning of robotic manipulation and robot force control.
\end{IEEEbiography}
\begin{IEEEbiography}[{\includegraphics[width=1in,height=1.25in,clip,keepaspectratio]{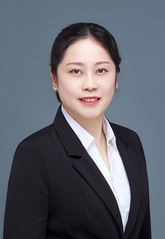}}]
{Fang Fang} received the B.S. degree in automation from Anhui University, Hefei, China in 2001, and the Ph.D. degree in control science and engineering from the Southeast University, Nanjing, China, in 2007.

She is currently an Associate Professor with the School of Automation, Southeast University. Her research interests include control and decision of service robot system, robot intelligent perception.
\end{IEEEbiography}
\begin{IEEEbiography}[{\includegraphics[width=1in,height=1.25in,clip,keepaspectratio]{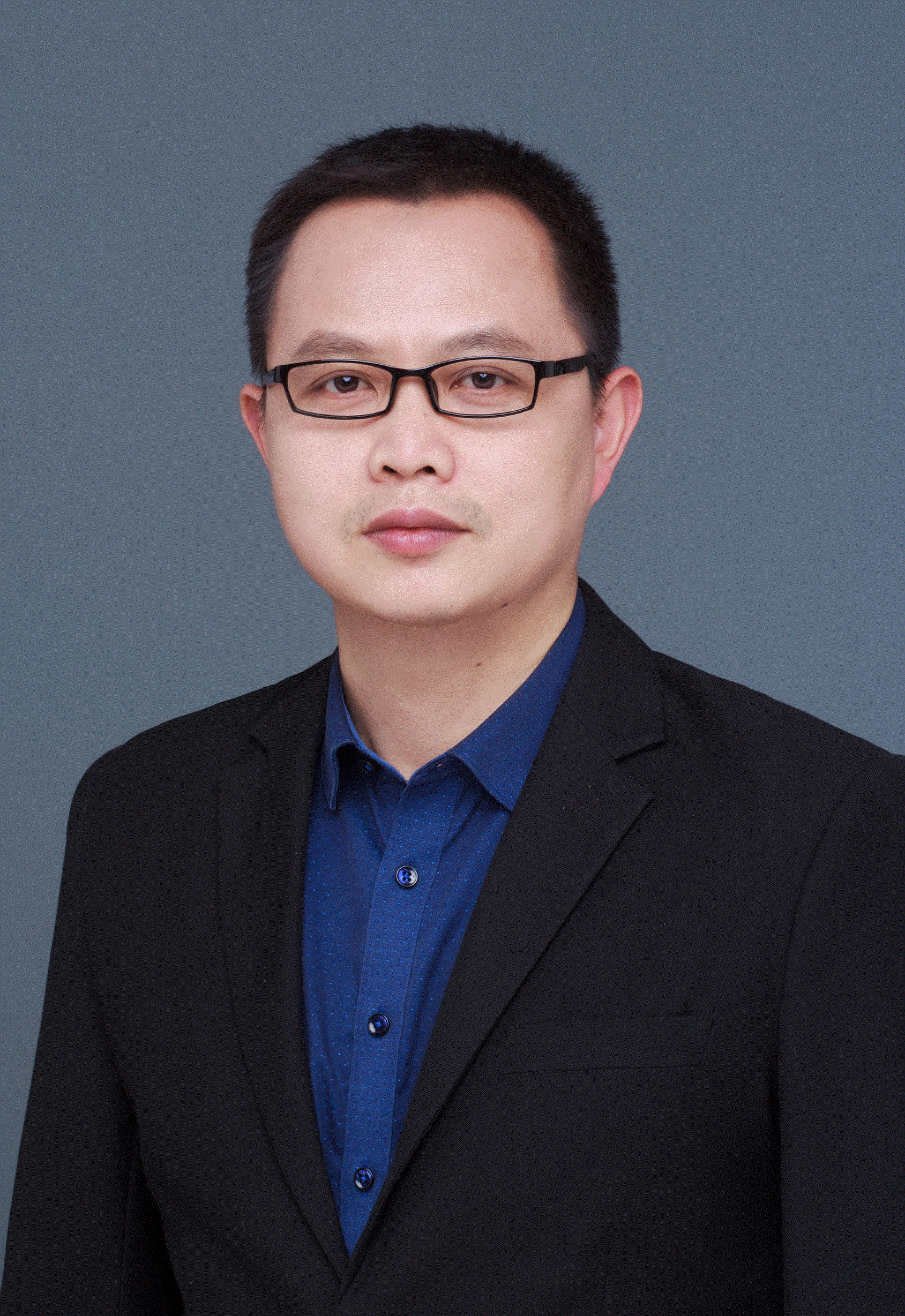}}]
{Shihua Li} received the B.S., M.S., and Ph.D. degrees in automatic control from Southeast University, Nanjing, China, in 1995, 1998, and 2001, respectively. Since 2001, he has been with the School of Automation, Southeast University, where he is currently a Full Professor and the Director of the Mechatronic Systems Control Laboratory.

His main research interests lie in modeling, analysis, and nonlinear control theory with applications to mechatronic systems, including manipulator, robot, AC motor, engine control, and power electronic systems.

\end{IEEEbiography}
\end{document}